\documentclass[10pt,twocolumn,letterpaper]{article}

\usepackage{cvpr}
\usepackage{times}
\usepackage{epsfig}
\usepackage{graphicx}
\usepackage{amsmath}
\usepackage{amssymb}
\usepackage{booktabs}
\usepackage{subfigure}
\usepackage{threeparttable}
\usepackage[boxed, ruled,linesnumbered]{algorithm2e}

\usepackage{amsmath,amsfonts,amssymb,amsthm}
\usepackage{mathtools}
\DeclarePairedDelimiter{\norm}{\lVert}{\rVert}
\DeclareMathOperator*{\argmin}{argmin} 

\usepackage[pagebackref=true,breaklinks=true,letterpaper=true,colorlinks,bookmarks=false]{hyperref}

\cvprfinalcopy 

\def\cvprPaperID{3083} 

\ifcvprfinal\fi
\begin{document}

\title{Contrastive Adaptation Network for Unsupervised Domain Adaptation}

\author{Guoliang Kang$^{1}$, Lu Jiang$^2$, 
Yi Yang$^{1,3}$\thanks{
Corresponding author. Part of this work was done when Yi Yang was visiting Baidu Research during his Professional Experience Program.}, 
Alexander G. Hauptmann$^4$\\
$^1$CAI, University of Technology Sydney, 
$^2$Google AI, 
$^3$Baidu Research, 
$^4$Carnegie Mellon University
\\
{\tt\small kgl.prml@gmail.com, }
{\tt\small lujiang@google.com, }
{\tt\small Yi.Yang@uts.edu.au, }
{\tt\small alex@cs.cmu.edu}
}

\newcommand{\lu}[1]{{\color{red}{(lu: #1)}}}
\newcommand{\kang}[1]{{\color{blue}{(kang: #1)}}}

\newcommand{\x}{\boldsymbol{x}}
\newcommand{\X}{\boldsymbol{X}}
\newcommand{\ibf}[1]{\textit{\textbf{#1}}}
\newcommand{\pl}{\phi_l}
\newcommand{\kl}{k_l}
\newcommand{\De}{\hat{\mathcal{D}}}
\newcommand{\D}{\mathcal{D}}

\maketitle

\begin{abstract}
Unsupervised Domain Adaptation (UDA) makes predictions for the target domain data while manual annotations are only available in the source domain. Previous methods minimize the domain discrepancy neglecting the class information, which may lead to misalignment and poor generalization performance. To address this issue, this paper proposes Contrastive Adaptation Network (CAN) optimizing a new metric which explicitly models the intra-class domain discrepancy and the inter-class domain discrepancy. We design an alternating update strategy for training CAN in an end-to-end manner. Experiments on two real-world benchmarks Office-31 and VisDA-2017 demonstrate that CAN performs favorably against the state-of-the-art methods and produces more discriminative features.
\end{abstract}

\section{Introduction}
Recent advancements in deep neural networks have successfully improved a variety of learning problems \cite{zhu2017bidirectional,dong2019search,Yawei2019Taking,jiang2014easy,kang2018shakeout}. For supervised learning, however, massive labeled training data is still the key to learning an accurate deep model. Although abundant labels may be available for a few pre-specified domains, such as ImageNet \cite{deng2009imagenet}, manual labels often turn out to be difficult or expensive to obtain for every ad-hoc target domain or task. The absence of in-domain labeled data hinders the application of data-fitting models in many real-world problems.

In the absence of labeled data from the target domain, Unsupervised Domain Adaptation (UDA) methods have emerged to mitigate the domain shift in data distributions \cite{ben2007analysis,ben2010theory,bruzzone2010domain,tzeng2017adversarial,saenko2010adapting,hoffman2016fcns,bousmalis2017unsupervised,hoffman2017cycada}. 
It relates to unsupervised learning as it requires manual labels only from the source domain and zero labels from the target domain.
Among the recent work on UDA, a seminal line of work proposed by Long \etal \cite{long2015learning,long2017deep} aims at minimizing the discrepancy between the source and target domain in the deep neural network, where the domain discrepancy is measured by Maximum Mean Discrepancy (MMD) \cite{long2015learning} and Joint MMD (JMMD) \cite{long2017deep}. 
MMD and JMMD have proven effective in many computer vision problems and demonstrated the state-of-the-art results on several UDA benchmarks~\cite{long2015learning,long2017deep}.

\begin{figure}[t]
\begin{center}
\includegraphics[scale=0.25]{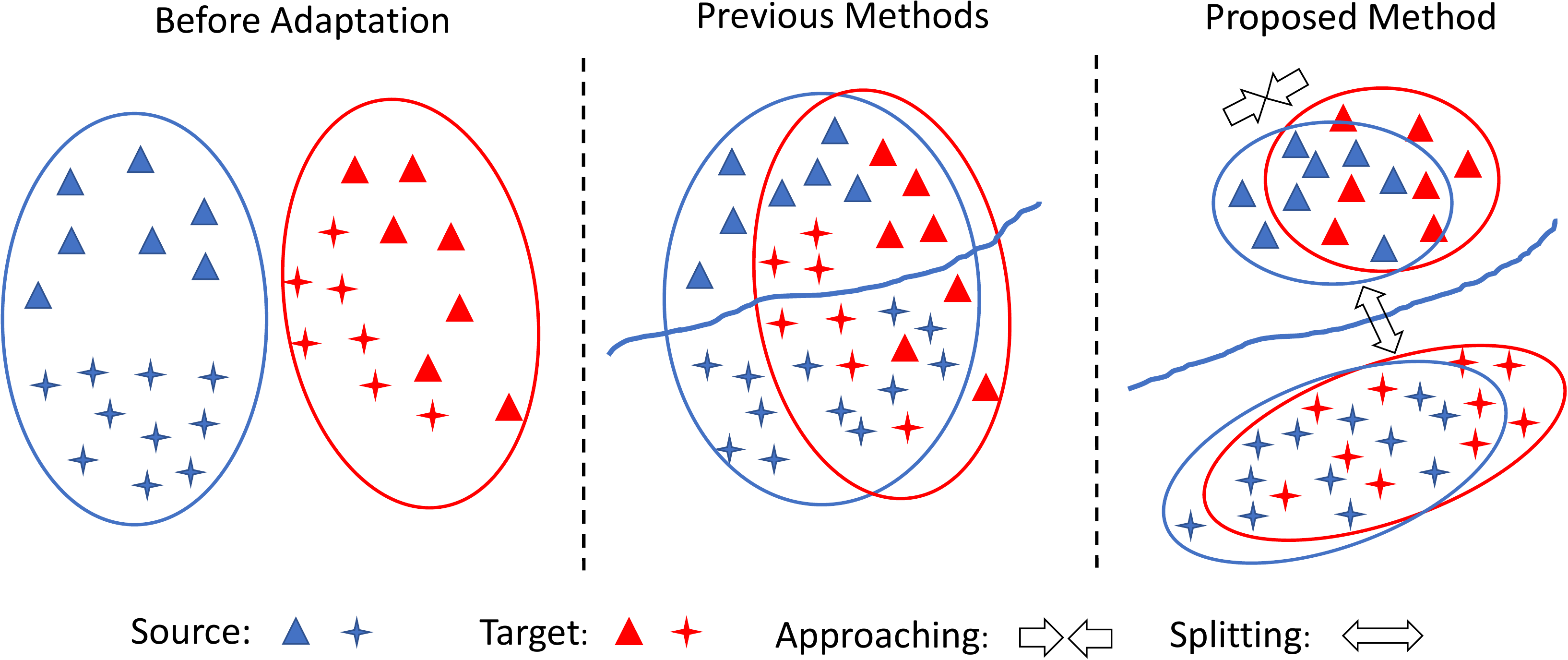}
\end{center}
\caption{\label{fig:motivation}
Comparison between previous domain-discrepancy minimization methods and ours.
\textbf{Left:} The domain shift exists between the source and target data before adaptation.
\textbf{Middle:}
Class-agnostic adaptation aligns source and target data at the domain-level, neglecting the class label of the sample, and hence may lead to sub-optimal solutions.
Consequently, the target samples of one label may be misaligned with source samples of a different label. 
\textbf{Right:} Our method performs class-aware alignment across domains. 
To avoid the misalignment, only the intra-class domain discrepancy is minimized. 
The inter-class domain discrepancy is maximized to enhance the model's generalization ability.
}
\end{figure}
Despite the success of previous methods based on MMD and JMMD, 
most of them measure the domain discrepancy at the \textit{domain} level, neglecting the class from which the samples are drawn. These class-agnostic approaches, hence, do not discriminate whether samples from two domains should be aligned according to their class labels 
(Fig.~\ref{fig:motivation}). 
This can impair the adaptation performance due to the following reasons.
First, samples of different classes may be aligned incorrectly, 
\eg both MMD and JMMD can be minimized even when the target-domain samples are misaligned with the source-domain samples of a different class. 
Second, the learned decision boundary may generalize poorly for the target domain. There exist many sub-optimal solutions near the decision boundary. 
These solutions may overfit the source data well but are less discriminative for the target. 

To address the above issues, we introduce a new \emph{Contrastive Domain Discrepancy (CDD)} objective to enable class-aware UDA. 
We propose to minimize the intra-class discrepancy, \ie the domain discrepancy within the same class, and maximize the inter-class margin, \ie the domain discrepancy between different classes.
Considering the toy example in Fig.~\ref{fig:motivation}, CDD will draw closer the source and target samples of the same underlying class (\eg the blue and red triangles), while 
pushing apart the samples from different classes (\eg the blue triangle and the red star).

Unfortunately, to estimate and optimize with CDD, we may not train a deep network out-of-the-box as we need to overcome the following two technical issues.
First, we need labels from both domains to compute CDD, however, target labels are unknown in UDA.
A straightforward way, of course, is to estimate the target labels by the network outputs during training.
However, because the estimation can be noisy, we find it can harm the adaptation performance (see Section~\ref{sec:ablation-study}).
Second, during the mini-batch training, 
for a class $C$, the mini-batch may only contain samples from one domain (source or target), 
rendering it infeasible to estimate the intra-class domain discrepancy of $C$. 
This can result in a less efficient adaptation.
The above issues require special design of the network and the training paradigm.

In this paper, we propose \emph{Contrastive Adaptation Network (CAN)} 
to facilitate the optimization with CDD. 
During training, in addition to minimizing the cross-entropy loss on labeled source data,
CAN alternatively estimates the underlying label hypothesis of target samples through clustering, 
and adapts the feature representations 
according to the CDD metric.
After clustering,
the ambiguous target data (\ie far from the cluster centers) 
and ambiguous classes (\ie containing few target samples around the cluster centers) 
are zeroed out in estimating the CDD.
Empirically we find that during training, 
an increasing amount of samples will be 
taken into account.
Such progressive learning can help CAN capture more accurate statistics of data distributions. 
Moreover, to facilitate the mini-batch training of CAN, 
we employ the class-aware sampling for both source and target domains,
\ie at each iteration, we sample data from both domains for each class within a randomly sampled class subset. Class-aware sampling can improve the training efficiency and the adaptation performance.

We validate our method on two public UDA benchmarks: 
Office-31 \cite{saenko2010adapting} and VisDA-2017 \cite{peng2017visda}. 
The experimental results show that our method performs favorably against the state-of-the-art UDA approaches, \ie we achieve the best-published result on the Office-31 benchmark and very competitive result on the challenging VisDA-2017 benchmark. Ablation studies are presented to verify the contribution of each key component in our framework.

In a nutshell, our contributions are as follows, 
\begin{itemize}
\item We introduce a new discrepancy metric \textit{Contrastive Domain Discrepancy} (CDD) to perform class-aware alignment for unsupervised domain adaptation.
\item We propose a network \textit{Contrastive Adaptation Network} to facilitate the end-to-end training with CDD.
\item Our method achieves the best-published result on the Office-31 benchmark~\cite{saenko2010adapting} and competitive performance 
compared to the state-of-the-art on the challenging VisDA-2017 benchmark~\cite{peng2017visda}.
\end{itemize}

\section{Related Work} \label{sec:related-work}

%
%

\noindent\textbf{Class-agnostic domain alignment.}
A common practice for UDA is to minimize the discrepancy between domains 
to obtain domain-invariant features ~\cite{ganin2014unsupervised,bousmalis2016domain,long2017deep,long2015learning,long2016unsupervised,sun2016deep,kang2018deep}. 
For example, Tzeng \etal \cite{tzeng2014deep} proposed a kind of domain confusion loss to encourage the network to learn both semantically meaningful and domain invariant representations. Long \etal  proposed DAN \cite{long2015learning} and JAN \cite{long2017deep} to minimize the MMD and Joint MMD distance across domains respectively, over the domain-specific layers.
Ganin \etal \cite{ganin2014unsupervised} enabled the network to learn domain invariant representations in adversarial way by 
back-propagating the reverse gradients of the domain classifier. 
Unlike these domain-discrepancy minimization methods, our method performs  \emph{class-aware} domain alignment.

\noindent\textbf{Discriminative domain-invariant feature learning.}
Some previous works pay efforts to learn more disciminative features while 
performing domain alignment \cite{sener2016learning,haeusser2017associative,saito2017adversarial,saito2017maximum,pei2018multi,wang2018visual}.
Adversarial Dropout Regularization (ADR) \cite{saito2017adversarial} 
and Maximum Classifier Discrepancy (MCD) \cite{saito2017maximum} were proposed to train 
a deep neural network in adversarial way to avoid generating non-discriminative features lying in the region near the decision boundary.
Similar to us, Long \etal \cite{long2013transfer} and Pei \etal \cite{pei2018multi} take the class information into account while measuring the domain discrepancy. 
However, our method differs from theirs mainly in two aspects. 
Firstly, we explicitly model two types of domain discrepancy, \ie
the intra-class domain discrepancy and the inter-class domain discrepancy.
The inter-class domain discrepancy, which has been ignored by most previous methods,
is proved to be beneficial for enhancing 
the model adaptation performance.
Secondly, in the context of deep neural networks, 
we treat the training process as 
an alternative optimization over target label hypothesis and features.

\noindent\textbf{Intra-class compactness and inter-class separability modeling.}
This paper is also related to the work that explicitly models the intra-class compactness and the inter-class separability, \eg the contrastive loss \cite{hadsell2006dimensionality} and the triplet loss \cite{schroff2015facenet}. These methods have been used in various applications, 
\eg face recognition \cite{cheng2016person}, person re-identification \cite{hermans2017defense}, etc. Different from these methods designed for a single domain, our work focuses on adaptation across domains. 
\section{Methodology}

Unsupervised Domain Adaptation (UDA) aims at improving the model's generalization performance on target domain
by mitigating the domain shift in data distribution of the source and target domain.
Formally, given a set of source domain samples 
$\mathcal{S} = \{(\boldsymbol{x}^s_1, y^s_1), \cdots, (\boldsymbol{x}^s_{N_s}, y^s_{N_s})\}$,
and target domain samples
$\mathcal{T} = \{\boldsymbol{x}^t_1, \cdots, \boldsymbol{x}^t_{N_t}\}$, $\boldsymbol{x}^s$, $\boldsymbol{x}^t$ represent the input data, 
and $y^s \in \{0, 1, \cdots, M-1\}$ denote the source data label of $M$ classes. 
The target data label ${y^t} \in \{0, 1, \cdots, M-1\}$ is unknown.
Thus, in UDA,  
we are interested in training a network using labeled source domain data $\mathcal{S}$ and unlabeled target domain data $\mathcal{T}$ to make accurate predictions $\{\hat{y}^t\}$ on $\mathcal{T}$.

We discuss our method in the context of deep neural networks.
In deep neural networks, a sample owns hierarchical features/representations denoted by the activations of each layer $l \in \mathcal{L}$. 
In the following, 
we use $\phi_{l}(\x)$ 
to denote the outputs of layer $l$ in a deep neural network  $\Phi_\theta$ for the input $\x$,
where $\phi(\cdot)$ denotes the mapping defined by the deep neural network 
from the input to a specific layer.

In the rest of this section, 
we start our discussions by briefly reviewing the relevant concepts in MMD in Section \ref{revisit}. Section~\ref{sec:cdd} introduces a new domain discrepancy metric.  Finally, Section~\ref{sec:can} and Section~\ref{sec:can_training} discuss the objective and 
the training procedure of proposed deep network.


\subsection{Maximum Mean Discrepancy Revisit} \label{revisit}

In Maximum Mean Discrepancy (MMD), \{$\boldsymbol{x}^s_i$\} and \{$\boldsymbol{x}^t_i$\} are \textit{i.i.d.} sampled from  the marginal distributions $P(\boldsymbol{X}^s)$ and $Q(\boldsymbol{X}^t)$ respectively. Based on the observed samples, 
MMD \cite{sejdinovic2013equivalence} performs a kernel two-sample test to determine whether to accept the null hypothesis $P = Q$ or not.
MMD is motivated by the fact that if two distributions are identical, all of their statistics should be the same. Formally, MMD defines the difference between two distributions with their mean embeddings 
in the reproducing kernel Hilbert space (RKHS), \ie
\begin{align}
\mathcal{D}_{\mathcal{H}}(P, Q) \triangleq \sup_{f\sim \mathcal{H}}{(\mathbb{E}_{\X^s}[f(\X^s)] - \mathbb{E}_{\X^t}[f(\X^t)])}_\mathcal{H},
\end{align}
where $\mathcal{H}$ is class of functions.

In practice, for a layer $l$, the squared value of MMD is estimated with the empirical kernel mean embeddings 
\begin{align}
\hat{\mathcal{D}}^{mmd}_l &= \frac{1}{n_{s}^2}\sum_{i=1}^{n_{s}}\sum_{j=1}^{n_{s}}{k_{l}(\phi_l(\x^{s}_i), \phi_l(\x^{s}_j))} \nonumber \\
					      &+ \frac{1}{n_{t}^2}\sum_{i=1}^{n_{t}}\sum_{j=1}^{n_{t}}{k_{l}(\phi_l(\x^{t}_i), \phi_l(\x^{t}_j))} \nonumber \\
					      &- \frac{2}{n_{s}n_{t}}\sum_{i=1}^{n_{s}}\sum_{j=1}^{n_{t}}{k_{l}(\phi_l(\x^{s}_i), \phi_l(\x^{t}_j))},
\end{align}
where $x^s\in \mathcal{S}' \subset \mathcal{S}$, $x^t\in \mathcal{T}' \subset \mathcal{T}$, 
$n_s =\lvert\mathcal{S}'\rvert$, $n_t =\lvert\mathcal{T}'\rvert$. 
The $\mathcal{S}'$ and $\mathcal{T}'$ represent the mini-batch source and target data sampled from 
$\mathcal{S}$ and $\mathcal{T}$ respectively.
And $k_{l}$ denotes the kernel selected for the $l$-th layer of deep neural network.

%

\subsection{Contrastive Domain Discrepancy}\label{sec:cdd}

We propose to explicitly take the class information into account and measure the \emph{intra-class} and \emph{inter-class} discrepancy across domains.
The intra-class domain discrepancy is minimized to compact the feature representations of  samples within a class,
whereas the inter-class domain discrepancy is maximized to push the representations of each other further away from the decision boundary. The intra-class and inter-class discrepancies are jointly optimized to improve the adaptation performance.

The proposed \textbf{C}ontrastive \textbf{D}omain \textbf{D}iscrepancy (CDD) is established on 
the difference between 
\textit{conditional} data distributions across domains.
Without any constraint on the type (\eg marginal or conditional) of data distributions, 
MMD is convenient to measure such difference between $P(\phi(\X^{s})|Y^s)$ and $Q(\phi(\X^{t})|Y^t)$, \ie 
$\mathcal{D}_{\mathcal{H}}(P, Q) \triangleq \sup_{f\sim \mathcal{H}}
{(\mathbb{E}_{\X^s}[f(\phi(\X^s)|Y^s)] - \mathbb{E}_{\X^t}[f(\phi(\X^t)|Y^t)])}_\mathcal{H}$.

Supposing 
$\mu_{cc'}(y, y') = \left\{ \begin{array}{ll}
         1 & \mbox{if $y=c, y'=c'$};\\
         0 & \mbox{otherwise}.\end{array} \right.
         $, 
for two classes $c_1, c_2$ (which can be same or different),
the kernel mean embedding estimation for squared $\mathcal{D}_{\mathcal{H}}(P, Q)$ is 
\begin{align}
\hat{\mathcal{D}}^{c_1c_2}(\hat{y}^t_1, \hat{y}^t_2, \cdots, \hat{y}^t_{n_t}, \phi) = e_1 + e_2 - 2e_3
\label{class-mmd}
\end{align}
where
\begin{align}
e_1 = \sum_{i=1}^{n_s}\sum_{j=1}^{n_s}{\frac{\mu_{c_1c_1}(y^s_i, y^s_j)k(\phi(\x^{s}_i), \phi(\x^{s}_j))}
{\sum_{i=1}^{n_s}\sum_{j=1}^{n_s}{\mu_{{c_1c_1}}(y^s_i, y^s_j)}}} \nonumber
\\
e_2 = \sum_{i=1}^{n_t}\sum_{j=1}^{n_t}{\frac{\mu_{c_2c_2}(\hat{y}^t_i, \hat{y}^t_j)k(\phi(\x^{t}_i), \phi(\x^{t}_j))}
{\sum_{i=1}^{n_t}\sum_{j=1}^{n_t}{\mu_{c_2c_2}(\hat{y}^t_i, \hat{y}^t_j)}}} \nonumber
\\
e_3 = \sum_{i=1}^{n_s}\sum_{j=1}^{n_t}{\frac{\mu_{c_1c_2}(y^s_i, \hat{y}^t_j)k(\phi(\x^{s}_i), \phi(\x^{t}_j))}
{\sum_{i=1}^{n_s}\sum_{j=1}^{n_t}{\mu_{c_1c_2}(y^s_i, \hat{y}^t_j)}}}.
\end{align}
%
%
Note that Eq. (\ref{class-mmd}) defines two kinds of class-aware domain discrepancy,
1) when $c_1 = c_2 = c$, it measures \textit{intra-class} domain discrepancy; 
2) when $c_1 \neq c_2$, it becomes the \textit{inter-class} domain discrepancy.
To compute the mask $\mu_{c_2c_2}(\hat{y}_i^t, \hat{y}_j^t)$ and
$\mu_{c_1c_2}(y_i^s, \hat{y}_j^t)$,
we need to estimate target labels $\{\hat{y}^t_i\}$, 
which will be discussed in Section \ref{sec:can_training}.

Based on the above definitions,
the CDD is calculated as 
(The $\hat{y}^t_1, \hat{y}^t_2, \cdots, \hat{y}^t_{n_t}$ 
is abbreviated as $\hat{y}^t_{1:n_t}$)
\vspace{-2mm}
\begin{align}
\label{cdd}
\hat{\mathcal{D}}^{cdd} &= \underbrace{\frac{1}{M}\sum_{c=1}^M \hat{\mathcal{D}}^{c c}(\hat{y}^t_{1:n_t}, \phi)}_{intra} \nonumber \\
& - \underbrace{\frac{1}{M(M-1)}\sum_{c=1}^M \sum_{\substack{c{'}=1 \\ c{'}\neq c}}^M \De^{cc'}(\hat{y}^t_{1:n_t}, \phi)}_{inter},
\end{align}
where the intra- and inter-class domain discrepancies will be optimized in the opposite direction.

Note although the estimation of the labels $\{\hat{y}^t_i\}$ can be noisy,
the CDD (which is established on MMD) in itself is robust the the noise to an extent.
Because MMD is determined by the mean embeddings of distributions in the RKHS, 
the sufficient statistics is less likely to be severely affected by the label noise, 
especially when the amount of data is large.
We will discuss and verify this in Section \ref{sec:ablation-study}.



\begin{figure*}[t]
\begin{center}
\includegraphics[scale=0.29]{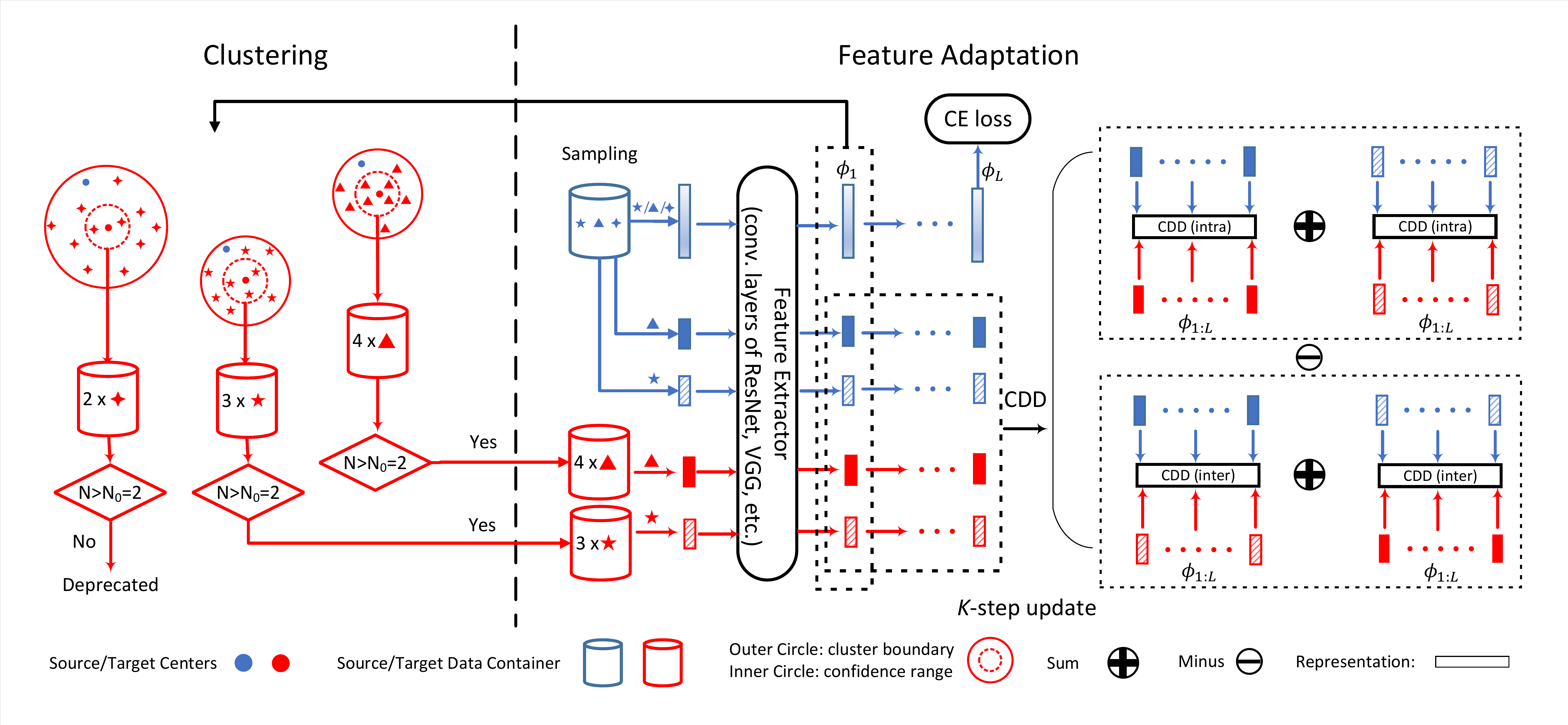}
\end{center}
\vspace{-3mm}
\caption{\label{fig:framework}
The training process of CAN. To minimize CDD, we perform alternative optimization 
between updating the target label hypothesis through clustering and 
adapting feature representations through back-propagation. 
For the clustering, we apply spherical K-means clustering of target samples
based on their current feature representations.
The number of clusters equal to that of underlying classes and 
the initial center of each class cluster is set to the center of source data
within the same class.
Then ambiguous data (\ie far from the affiliated cluster centers) and 
ambiguous classes (\ie containing few target samples around affiliated cluster centers) 
are discarded.
For the feature adaptation, the labeled target samples provided by the clustering stage
, together with the labeled source samples, pass through the network
to achieve their multi-layer feature representations. 
The features of domain-specific FC layers are adopted to estimate CDD (Eq. (\ref{cdd})).
Besides, we apply cross-entropy loss on independently sampled source data.
Back-propagating with minimizing CDD and cross-entropy loss (Eq. (\ref{obj-full})) adapts the
features and provides class-aware alignment.
Detailed descriptions can be found in Section \ref{sec:can_training}.}
\vspace{-3mm}
\end{figure*}
\subsection{Contrastive Adaptation Network}\label{sec:can}
Deep convolutional neural networks (CNNs) is able to learn more transferable features than shallow methods. 
However, the discrepancy still exists for domain-specific layers.
Specifically, the convolutional layers extracting general features are more transferable, 
while the fully-connected (FC) layers which exhibit abstract and domain-specific features should be adapted \cite{long2015learning,long2017deep}. 

In this paper, we start from ImageNet \cite{deng2009imagenet} pretrained networks, \eg ResNet \cite{he2016deep,he2016identity},
and replace the last FC layer with task-specific ones.
We follow the general practice that minimizes the domain discrepancy of last FC layers 
and fine-tunes the convolutional layers through back-propagation. 
Then our proposed CDD can be readily incorporated into the objective as  
an adaptation module
over the activations of FC layers.
We name our network \textbf{C}ontrastive \textbf{A}daptation \textbf{N}etwork (CAN).

\textbf{The overall objective.}
In a deep CNN,
we need to minimize CDD over multiple FC layers, \ie
minimizing
\begin{align}
\De^{cdd}_\mathcal{L} = \sum_{l=1}^{L}{\De_l^{cdd}}.
\label{cdd-l}
\end{align}

Besides,
we train the network with labeled source data through minimizing the cross-entropy loss,
\begin{align}
\ell^{ce} &= -\frac{1}{n{'}_s}\sum_{i{'}=1}^{n{'}_s}{\log{P_\theta(y^s_{i{'}}|\x^s_{i{'}})}}
\label{ce-loss}
\end{align}
where $y^s \in \{0, 1, \cdots, M-1\} $ is the ground-truth label of sample $\x^s$.
$P_\theta(y|\x)$ denotes the predicted probability of label ${y}$ with the network 
parameterized by $\theta$, given input $\x$.

Therefore, the overall objective can be formulated as  
\begin{align}
\min_\theta \ell &= \ell^{ce} + \beta \hat{\mathcal{D}}^{cdd}_\mathcal{L} \label{obj-full}
\end{align}
where 
$\beta$ is 
the weight of the discrepancy penalty term.
Through minimizing $\hat{\mathcal{D}}^{cdd}_\mathcal{L}$,
the intra-class domain discrepancy is minimized and the inter-class domain discrepancy is maximized
to perform class-aware domain alignment. 

Note that we independently sample the labeled source data 
to minimize the cross-entropy loss $\ell^{ce}$
and those to estimate the CDD $\hat{\mathcal{D}}^{cdd}_\mathcal{L}$.
In this way, we are able to design more efficient sampling strategy (see Section \ref{sec:can_training}) to facilitate the mini-batch stochastic optimization with CDD,
while not disturbing the conventional optimization with cross-entropy loss on labeled source data.




%
%
%
%

\subsection{Optimizing CAN} \label{sec:can_training}
The framework of CAN is illustrated in Fig. \ref{fig:framework}.
In this section, we mainly focus on discussing how to minimize CDD loss in CAN.

\textbf{Alternative optimization (AO).}
As shown in Eq. (\ref{cdd}),
we need to jointly optimize the target label hypothesis $\hat{y}^t_{1:n_t}$
and the feature representations $\phi_{1:L}$.
We adopt alternative steps to perform such optimization.
In detail, at each loop, 
given current feature representations, \ie fixing $\theta$, 
we update target labels through clustering.
Then, based on the updated target labels $\hat{y}^t$, 
we estimate and minimize CDD to adapt the features, 
\ie update $\theta$ through back-propagation.

We employ the input activations $\phi_1(\cdot)$ of the first task-specific layer 
to represent a sample.
For example, in ResNet, each sample can be  
represented as the outputs of the global average pooling layer,
which are also the inputs of the following task-specific layer.
Then the spherical K-means is adopted to perform the clustering of target samples and 
attach corresponding labels. 
The number of clusters is the same as the number of underlying classes $M$. 
For each class,
the target cluster center $O^{tc}$ is initialized as the  
source cluster center $O^{sc}$, \ie
$O^{tc} \gets O^{sc}$, 
where
$O^{sc} = \sum_{i=1}^{N_{s}}\mathbf{1}_{y^s_{i}=c}
\frac{\phi_{1}(\x^s_{i})}{\norm{\phi_{1}(\x^s_{i})}}$,
$\mathbf{1}_{y_i^s=c}
\left\{ \begin{array}{ll}
         1 & \mbox{if $y_i^s=c$}\\
         0 & \mbox{otherwise}\end{array} \right. $
and $c=\{0, 1, \cdots, M-1\}$. 
For the metric measuring the distance between points $\boldsymbol{a}$ and $\boldsymbol{b}$ 
in the feature space,
we apply the cosine dissimilarity, \ie
$dist(\boldsymbol{a}, \boldsymbol{b}) = 
\frac{1}{2}(1 - \frac{\langle \boldsymbol{a}, \boldsymbol{b} \rangle}
{\norm{\boldsymbol{a}}\norm{\boldsymbol{b}}}$).

Then the clustering process is iteratively 
1) attaching labels for each target samples:
$\hat{y}^t_i \gets \argmin_c {dist(\phi_{1}(\x^t_i), O^{tc}})$,
and
2) updating the cluster centers: 
$O^{tc} \gets \sum_{i=1}^{N_{t}}{\mathbf{1}_{\hat{y}^t_{i}=c} \frac{\phi_{1}(\x^t_{i})}{\norm{\phi_{1}(\x^t_{i})}}}$,
till convergence or reaching the maximum clustering steps.

After clustering, each target sample $\x_{i}^t$ is assigned a label $\hat{y}_i^t$ same as its affiliated clusters.
Moreover, ambiguous data, which is far from its affiliated cluster center, is discarded,
\ie we select a subset
$\mathcal{\tilde{T}} = \{
(\x^t,\hat{y}^t) | dist(\phi_{1}(\x^t), O^{t(\hat{y}^t)}) < D_0, \x^t\in\mathcal{T}\}
\label{eq:sample-filtering}
$,
where $D_0\in [0, 1]$ is a constant.

Moreover, to give a more accurate estimation of the distribution statistics, we assume that the minimum number of samples in $\mathcal{\tilde{T}}$ assigned to each class, should be guaranteed.
The class which doesn't satisfy such condition will not be considered in current loop, \ie
at loop $T_e$, the selected subset of classes 
$\mathcal{C}_{T_e} = \{c | \sum_i^{\lvert \mathcal{\tilde{T}}\rvert}{\mathbf{1}_{\hat{y}_i^t=c}} > N_0, c\in \{0, 1, \cdots, M-1\}\}
\label{eq:class-filtering}
$,
where $N_0$ is a constant.

At the start of training, due to the domain shift, 
it is more likely to exclude partial classes.
However, as training proceeds,   
more and more classes are included.
The reason is two folds: 
1) as training proceeds, the model becomes more accurate 
and 
2) benefiting from the CDD penalty, 
the intra-class domain discrepancy becomes smaller, 
and the inter-class domain discrepancy becomes larger, 
so that the hard (\ie 
ambiguous) classes are able to be taken into account.



\textbf{Class-aware Sampling (CAS).}
In the conventional training of deep neural networks, 
a mini-batch of data is usually sampled at each iteration 
without being differentiated by their classes.
However, it will be less efficient for computing the CDD.
For example, for class $C$, 
there may only exist samples from one domain (source or target) in the mini-batch, 
thus the intra-class discrepancy could not be estimated.

We propose to use class-aware sampling strategy to 
enable the efficient update of network with CDD.
It is easy to implement. 
We randomly select a subset of classes $\mathcal{C}^{'}_{T_e}$ from $\mathcal{C}_{T_e}$,
and then sample source data and target data for each class in $\mathcal{C}^{'}_{T_e}$.
Consequently, in each mini-batch of data during training, 
we are able to estimate the intra-class discrepancy for each selected class.

\textbf{Algorithm.}
Algorithm \ref{alg} shows one loop of the AO procedure, 
\emph{i.e.} alternating between a clustering phase (Step 1-4), and a ${K}$-step network update phase (Step 5-11).  
The loop of AO is repeated multiple times in our experiments. 
Because the feature adapting process is relatively slower,
we asynchronously update the target labels and the network parameters
to make the training more stable and efficient.

\begin{algorithm}[h]
 \KwIn{\\ 
 source data: $\mathcal{S} = \{(\boldsymbol{x}^s_1, y^s_1), \cdots, (\boldsymbol{x}^s_{N_s}, y^s_{N_s})\}$, 

 target data: $\mathcal{T} = \{\boldsymbol{x}^t_1, \cdots, \boldsymbol{x}^t_{N_t}\}$
 
 
 \textbf{Procedure:}
 }
 

 %
 
  
  Forward $\mathcal{S}$ and compute the $M$ cluster centers $O^{sc}$ \; 
  
  Initialize $O^{tc}$: $O^{tc} \gets O^{sc}$ \;
  
  Cluster target samples $\mathcal{T}$ using spherical K-means\;
  
  Filter the ambiguous target samples and classes\;
  
  
  
 \For{($k \gets 1;k \le K;k \gets k + 1$)}{ 
  Class-aware sampling based on $\mathcal{C}^{'}_{T_e}, 
  \tilde{\mathcal{T}}$, and
  $\mathcal{S}$\;
  
  Compute $\hat{\mathcal{D}}^{cdd}_\mathcal{L}$ using Eq. (\ref{cdd-l})\;
  
  Sample from $\mathcal{S}$ and compute $\ell^{ce}$ using Eq. (\ref{ce-loss})\;
  
  Back-propagate with the objective $\ell$ (Eq.(\ref{obj-full}))\;
  
  Update network parameters $\theta$.
 }
 \caption{Optimization of CAN at loop $T_e$.}
 \label{alg}
\end{algorithm}

\section{Experiments}\label{sec:experiments}
\subsection{Setups}
\noindent\textbf{Datasets:} We validate our method on two public benchmarks. \emph{Office-31} \cite{saenko2010adapting} is a common dataset for real-world domain adaptation tasks. 
It consists of 4,110 images belonging to 31 classes. 
This dataset contains three distinct domains, \emph{i.e.}, images which are collected from the 1) Amazon website (\textbf{A}mazon domain), 
2) web camera (\textbf{W}ebcam domain), 
and 3) digital SLR camera (\textbf{D}SLR domain) under different settings, respectively.
The dataset is imbalanced across domains, 
with 2,817 images in \textbf{A} domain, 795 images in \textbf{W} domain, and 498 images in \textbf{D} domain.

\begin{figure}[t]
\includegraphics[scale=0.25]{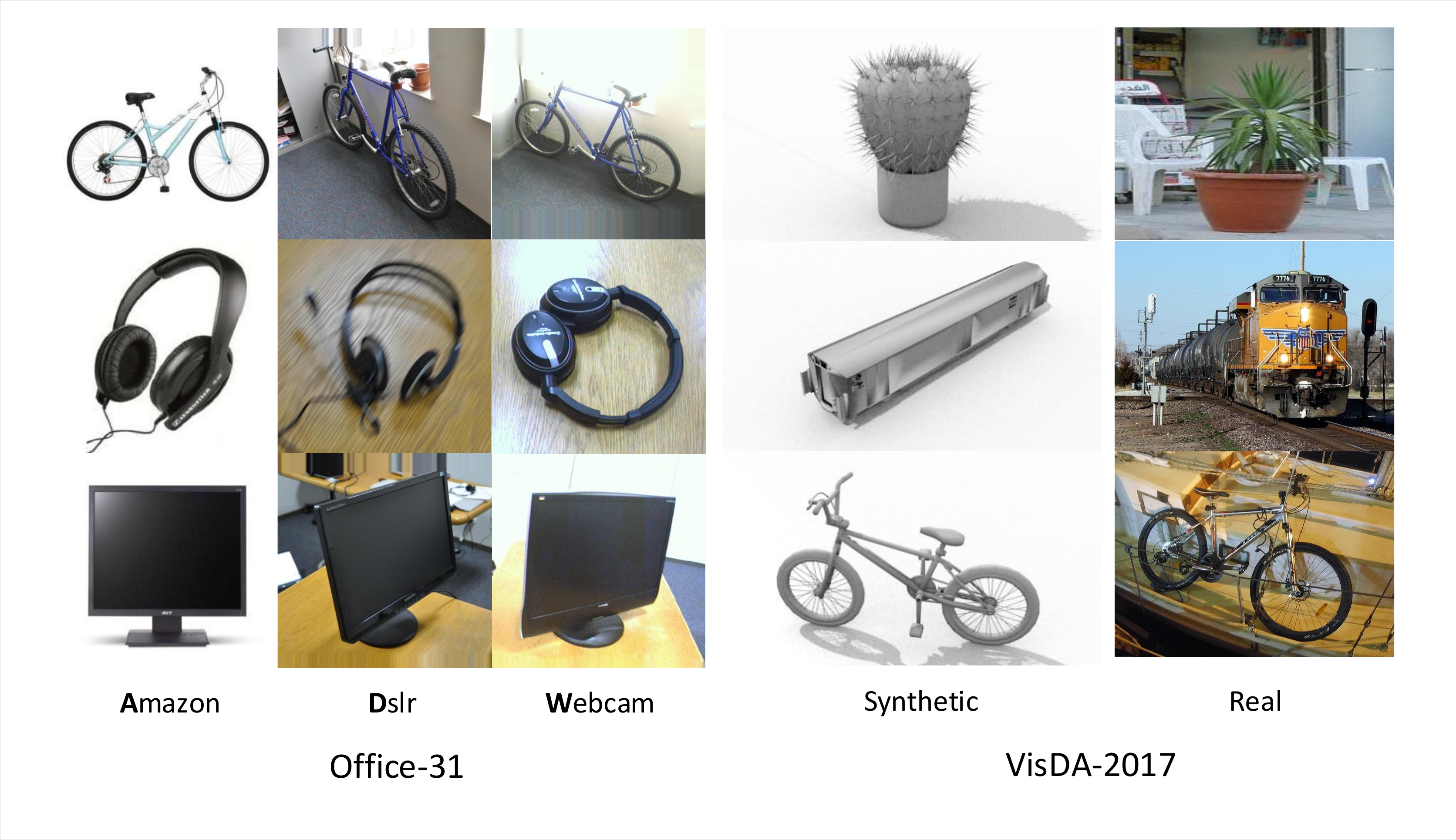}
\caption{\label{fig:gallery}
The gallery of Office-31 and VisDA-2017 datasets.
}
\vspace{-5mm}
\end{figure}

\noindent\emph{VisDA-2017} \cite{peng2017visda} is a challenging testbed for UDA 
with the domain shift from synthetic data to real imagery.
In this paper, we validate our method on its classification task.
In total there are $\sim$280k images from 12 categories.
The images are split into three sets, \ie a training set with 152,397 synthetic images,
a validation set with 55,388 real-world images,
and a test set with 72,372 real-world images.
The gallery of two datasets is shown in Fig. \ref{fig:gallery}

\noindent \textbf{Baselines:}
We compare our method with class-agnostic discrepancy minimization methods:
RevGrad \cite{ganin2014unsupervised,ganin2016domain}, DAN \cite{long2015learning}, and 
JAN \cite{long2017deep}. Moreover, we compare our method with the ones which explicitly or implicitly 
take the class information or decision boundary into consideration to learn 
more discriminative features: MADA \cite{pei2018multi}, MCD \cite{saito2017maximum}, 
and ADR \cite{saito2017adversarial}.The descriptions of these methods can be found in 
Section \ref{sec:related-work}.
We implement DAN and JAN using the released code \footnote{\url{https://github.com/thuml/Xlearn}}.
For a comparison under optimal parameter setting, we cite the performance of MADA, RevGrad, MCD and ADR reported in their corresponding papers~\cite{pei2018multi,saito2017adversarial,saito2017maximum,ganin2014unsupervised}.
\noindent\textbf{Implementation details:} We use ResNet-50 and ResNet-101 \cite{he2016deep,he2016identity} pretrained on ImageNet \cite{deng2009imagenet} as our backbone networks. 
We replace the last FC layer with the task-specific FC layer,
and finetune the model with labeled source domain data and unlabeled target domain data.
All the network parameters are shared between the source domain and target domain data
other than those of the batch normalization layers which are domain-specific.
The hyper-parameters are selected following the same protocol as described in \cite{long2015learning}, 
\ie we train a domain classifier and perform selection on a validation set (of labeled source samples and unlabeled target samples)
by jointly evaluating the test errors of the source classifier and the domain classifier. 

We use mini-batch stochastic gradient descent (SGD) 
with momentum of 0.9 to train the network. 
We follow the same learning rate schedule as described in \cite{ganin2014unsupervised,long2015learning,long2017deep}, 
\ie the learning rate $\eta_p$ is adjusted following $\eta_p = \frac{\eta_0}{(1+ap)^b}$, 
where $p$ linearly increases from 0 to 1. The $\eta_0$ is the initial learning rate,
\ie 0.001 for the convolutional layers and 0.01 for the task-specific FC layer.  
For Office-31, $a = 10$ and $b = 0.75$,
while for VisDA-2017, $a = 10$ and $b = 2.25$.
The $\beta$ selected is 0.3.
The thresholds $(D_0, N_0)$ are set to (0.05, 3) for Office-31 tasks 
\textbf{A}$\rightarrow$\textbf{W} and \textbf{A}$\rightarrow$\textbf{D}.
And we don't filter target samples and classes for other tasks during training.

\setlength{\tabcolsep}{4pt}
\begin{table*}[ht]
\small
\begin{center}
\scalebox{0.9}{
\begin{tabular}{ l  c  c   c   c   c   c  c}
\toprule
Method &  A $\rightarrow$ W & D $\rightarrow$ W & W $\rightarrow$ D & A $\rightarrow$ D & D $\rightarrow$ A & W $\rightarrow$ A & Average \\
\midrule
Source-finetune & 68.4 $\pm$ 0.2 & 96.7 $\pm$ 0.1 & 99.3 $\pm$ 0.1 & 68.9 $\pm$ 0.2 & 62.5 $\pm$ 0.3 & 60.7 $\pm$ 0.3 & 76.1 \\
RevGrad \cite{ganin2014unsupervised,ganin2016domain} & 82.0 $\pm$ 0.4 & 96.9 $\pm$ 0.2 & 99.1 $\pm$ 0.1 & 79.7 $\pm$ 0.4 & 68.2 $\pm$ 0.4 & 67.4 $\pm$ 0.5 & 82.2 \\ 
DAN \cite{long2015learning}   & 80.5 $\pm$ 0.4 & 97.1 $\pm$ 0.2 & 99.6 $\pm$ 0.1 & 78.6 $\pm$ 0.2 & 63.6 $\pm$ 0.3 & 62.8 $\pm$ 0.2 & 80.4 \\
JAN \cite{long2017deep}   & 85.4 $\pm$ 0.3 & 97.4 $\pm$ 0.2 & 99.8 $\pm$ 0.2 & 84.7 $\pm$ 0.3 & 68.6 $\pm$ 0.3 & 70.0 $\pm$ 0.4 & 84.3 \\
MADA \cite{pei2018multi}    & 90.0 $\pm$ 0.2 & 97.4 $\pm$ 0.1 & 99.6 $\pm$ 0.1 & 87.8 $\pm$ 0.2  & 70.3 $\pm$  0.3 & 66.4 $\pm$ 0.3 & 85.2 \\
\midrule
Ours (intra only) & 93.2 $\pm$ 0.2  &  98.4 $\pm$ 0.2   &  99.8 $\pm$ 0.2 &  92.9 $\pm$ 0.2 & 76.5 $\pm$ 0.3 & 76.0 $\pm$ 0.3 & 89.5      \\
Ours (CAN)        &  \textbf{94.5 $\pm$ 0.3}  & \textbf{99.1 $\pm$ 0.2} & \textbf{99.8 $\pm$ 0.2} &  \textbf{95.0 $\pm$ 0.3} & \textbf{78.0  $\pm$  0.3} & \textbf{77.0 $\pm$ 0.3}  & \textbf{90.6}   \\

\bottomrule
\end{tabular}}
\end{center}
\caption{Classification accuracy (\%) for all the six tasks of Office-31 dataset based on ResNet-50 \cite{he2016deep,he2016identity}.
Our methods named ``intra only'' and ``CAN'' 
are trained with intra-class domain discrepancy and contrastive domain discrepancy, respectively.
}
\label{table:cls-office-31}
\end{table*}

\begin{table*}[ht]
\small 
\begin{center}
\scalebox{0.9}{
\begin{tabular}{ l  c  c   c   c   c   c  c c c c c c c}
\toprule
Method & \rotatebox{90}{airplane} & \rotatebox{90}{bicycle} & \rotatebox{90}{bus} & \rotatebox{90}{car} & \rotatebox{90}{horse} & \rotatebox{90}{knife} & \rotatebox{90}{motorcycle} & \rotatebox{90}{person} & \rotatebox{90}{plant} & \rotatebox{90}{skateboard} & \rotatebox{90}{train} & \rotatebox{90}{truck} & Average \\
\midrule
Source-finetune   &  72.3 & 6.1 & 63.4 & \textbf{91.7} & 52.7 & 7.9 & 80.1 & 5.6  & 90.1 & 18.5 & 78.1 & 25.9 & 49.4 \\
RevGrad \cite{ganin2014unsupervised,ganin2016domain} & 81.9 & 77.7 & 82.8 & 44.3 & 81.2 & 29.5 & 65.1 & 28.6 & 51.9 & 54.6 & 82.8 & 7.8 & 57.4 \\
DAN \cite{long2015learning}   & 68.1 & 15.4 & 76.5 & 87.0 & 71.1 & 48.9 & 82.3 & 51.5 & 88.7 & 33.2 & 88.9 & 42.2 & 62.8 \\
JAN \cite{long2017deep}       & 75.7 & 18.7 & 82.3 & 86.3 & 70.2 & 56.9 & 80.5 & 53.8 & 92.5 & 32.2 & 84.5 & 54.5 & 65.7 \\
MCD \cite{saito2017maximum}   & 87.0 & 60.9 & 83.7 & 64.0 & 88.9 & 79.6 & 84.7 & 76.9 & 88.6 & 40.3 & 83.0 & 25.8 & 71.9 \\
ADR \cite{saito2017adversarial}   & 87.8 & 79.5 & 83.7 & 65.3 & 92.3 & 61.8 & 88.9 & 73.2 & 87.8 & 60.0 & 85.5 & 32.3 & 74.8 \\
SE \cite{french2017self}  &  95.9 & \textbf{87.4}   & \textbf{85.2} &  58.6  & 96.2  &  95.7  & 90.6 & 80.0 & 94.8 & 90.8 & 88.4 & 47.9 & 84.3 \\
\midrule
Ours (intra only)     & 96.5 & 72.1 & 80.9 & 70.8 & 94.6 & \textbf{98.0} & \textbf{91.7} & \textbf{84.2} & 90.3 & 89.8 & \textbf{89.4} & 47.9 & 83.9 \\
Ours (CAN)     & \textbf{97.0} & 87.2 & 82.5 & 74.3 & \textbf{97.8} & 96.2 & 90.8 & 80.7 & \textbf{96.6} & \textbf{96.3} & 87.5 & \textbf{59.9} & \textbf{87.2} \\
\bottomrule
\end{tabular}}
\end{center}
\caption{Classification accuracy (\%) on the VisDA-2017 validation set based on ResNet-101 \cite{he2016deep,he2016identity}. 
Our methods named ``intra only'' and ``CAN'' 
are trained with intra-class domain discrepancy and contrastive domain discrepancy, respectively.
}
\vspace{-3mm}
\label{table:cls-visda}
\end{table*}

\subsection{Comparison with the state-of-the-art}\label{sec:soa}
Table \ref{table:cls-office-31} shows the classification accuracy on six tasks of Office-31. 
All domain adaptation methods yield notable improvement over the ResNet model (first row) which is fine-tuned on labeled source data only. CAN outperforms other baseline methods across all tasks, achieving the state-of-the-art performance. On average, it boosts the accuracy of JAN by a absolute 6.3\% and that of MADA by 5.4\%.

\begin{figure}[t]
\begin{center}
\includegraphics[scale=0.20]{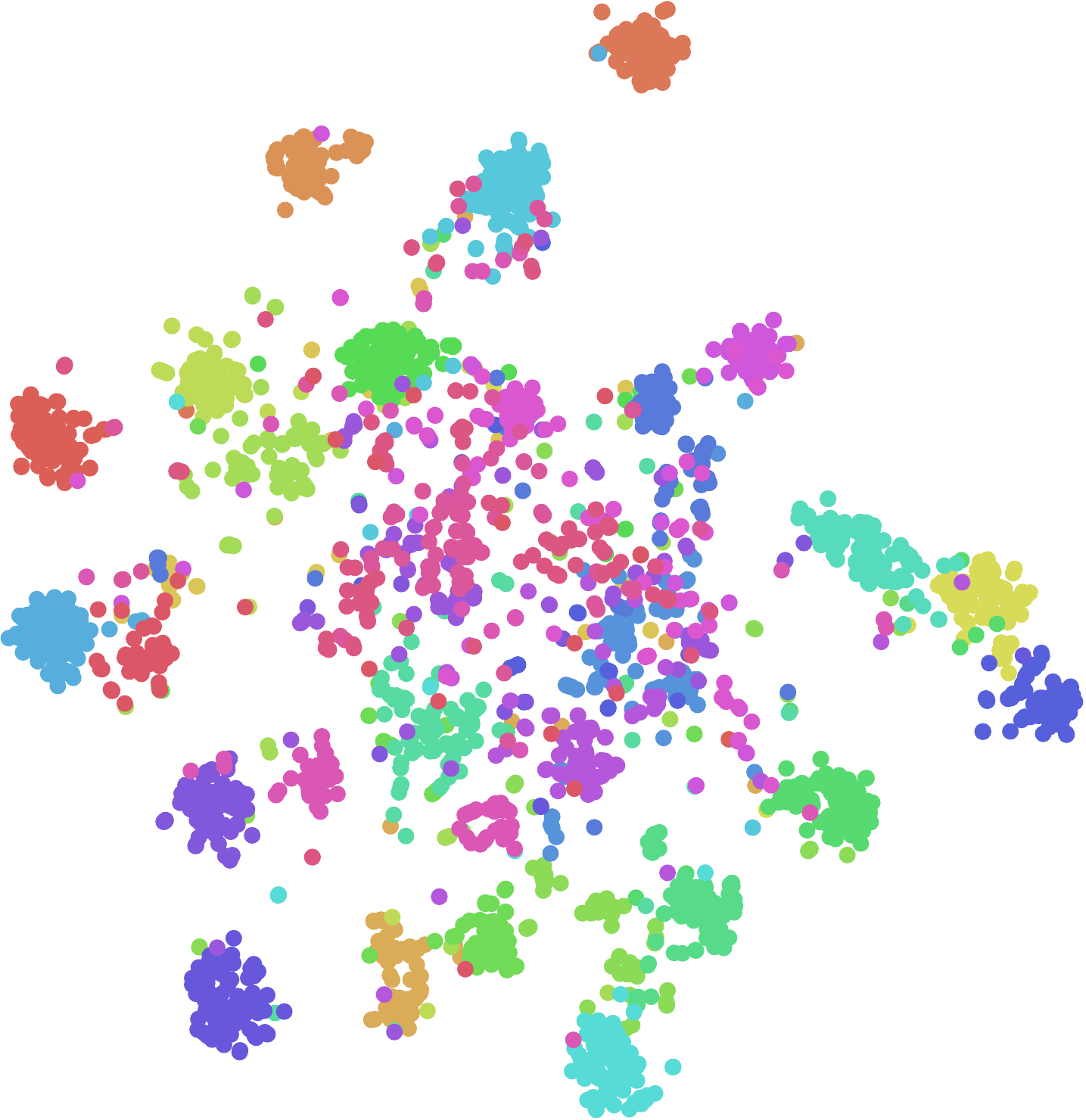}
\qquad
\includegraphics[scale=0.20]{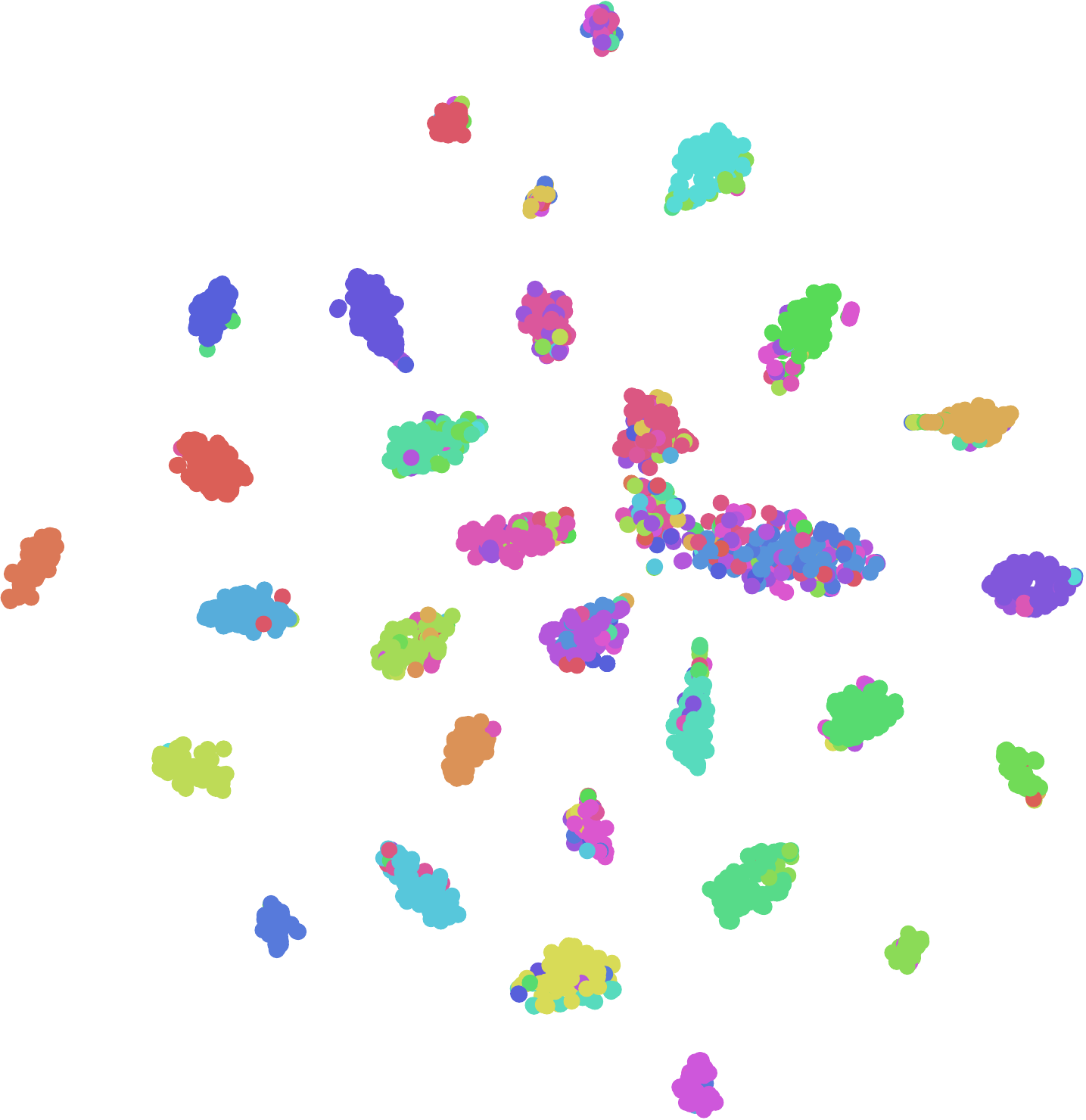}
\end{center}
\caption{\label{fig:tsne}
Visualization with t-SNE for different adaptation methods (bested viewed in color). \textbf{Left}: t-SNE of JAN. \textbf{Right}: CAN. The input activations of the last FC layer are used for the computation of t-SNE. The results are on Office-31 task $\textbf{W} \rightarrow \textbf{A}$.
}
\vspace{-4mm}
\end{figure}

We visualize the distribution of learned features by t-SNE \cite{maaten2008visualizing}. Fig.~\ref{fig:tsne} illustrates a representative task $\textbf{W} \rightarrow \textbf{A}$.
Compared to JAN,
as expected, the target data representations learned by CAN
demonstrate higher intra-class compactness 
and much larger inter-class margin. 
This suggests that our CDD produces more discriminative features for the target domain and substantiates our improvement in Table~\ref{table:cls-office-31}. 

Table \ref{table:cls-visda} lists the accuracy over 12 classes on VisDA-2017 with the validation set as the target domain. 
Our method outperforms the other baseline methods.
The mean accuracy of our model (87.2\%) 
outperforms
the self-ensembling (SE) method~\cite{french2017self} (84.3\%)
which wins the first place in the VisDA-2017 competition, by 2.9\%.
It is worth noting that SE mainly deals with UDA by ensemble and data augmentation,
which is orthogonal to the topic of this paper and thus can be easily combined to boost the performance further.

Moreover, we also perform adaptation on the VisDA-2017 test set (as the target
domain), and submit our predictions to official evaluation server. Our goal is to evaluate the effectiveness of our proposed technique based on a vanilla backbone (ResNet-101). We choose not to use ensemble or additional data augmentation which is commonly used to boost the performance in the competition.
Anyhow, our single model achieves a very competitive accuracy of 87.4\%, which 
is comparable to the method which ranks at the second place on the leaderboard (87.7\%).



From Table \ref{table:cls-office-31} and \ref{table:cls-visda}, we have two observations:
1) Taking class information/decision boundary into account is beneficial for the adaptation.
It can be seen that MADA, MCD, ADR and our method achieve
better performance than 
class-agnostic methods, 
\eg RevGrad, DAN, JAN, \etc.
2) Our way of exploiting class information is more effective. 
We achieve better accuracy than MADA (+5.4\%), ADR (+12.4\%), and MCD (+15.3\%).

\begin{figure*}[t]
\begin{center}
\par
\subfigure[]{
\includegraphics[scale=0.29]{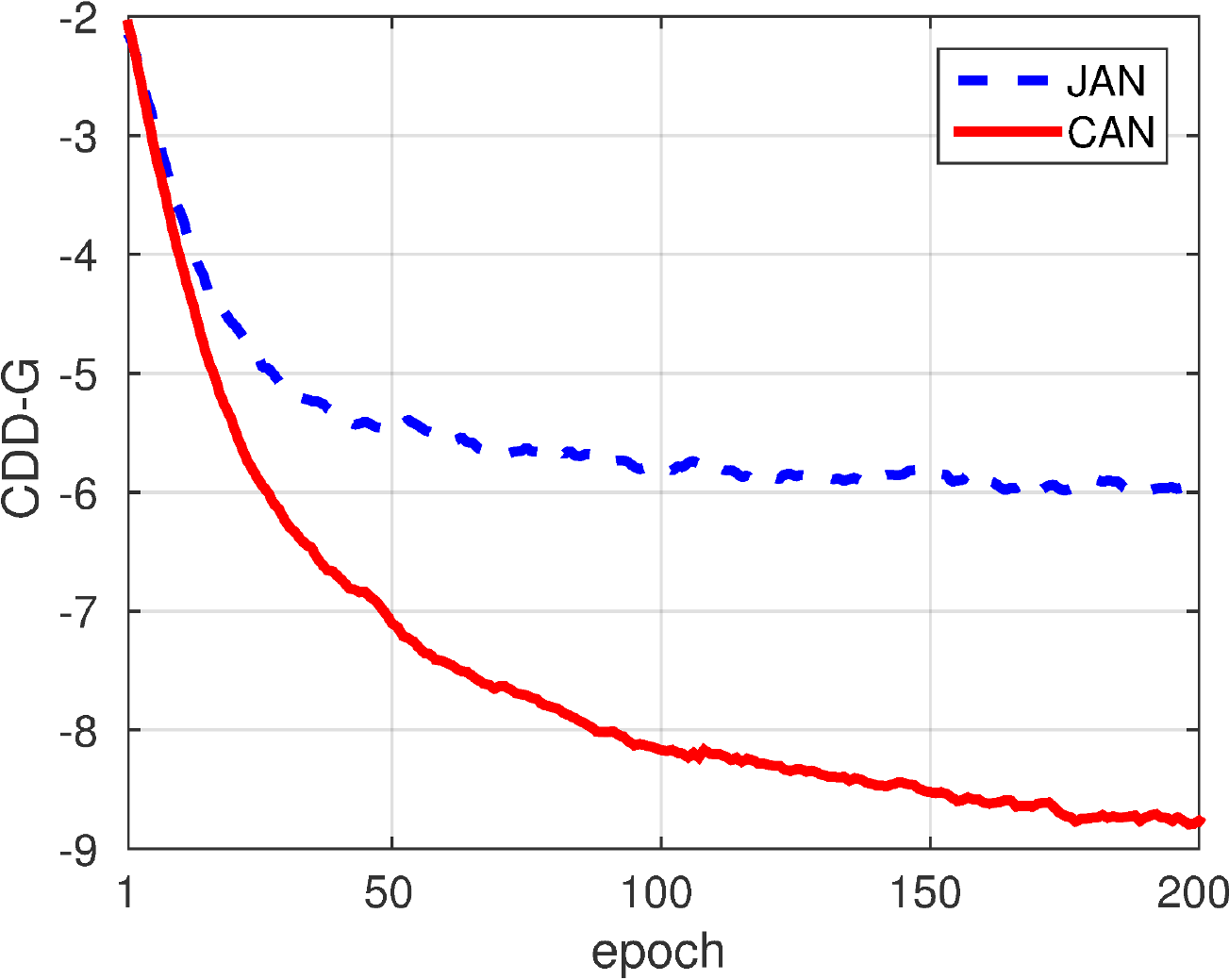}
}
\subfigure[]{
\includegraphics[scale=0.29]{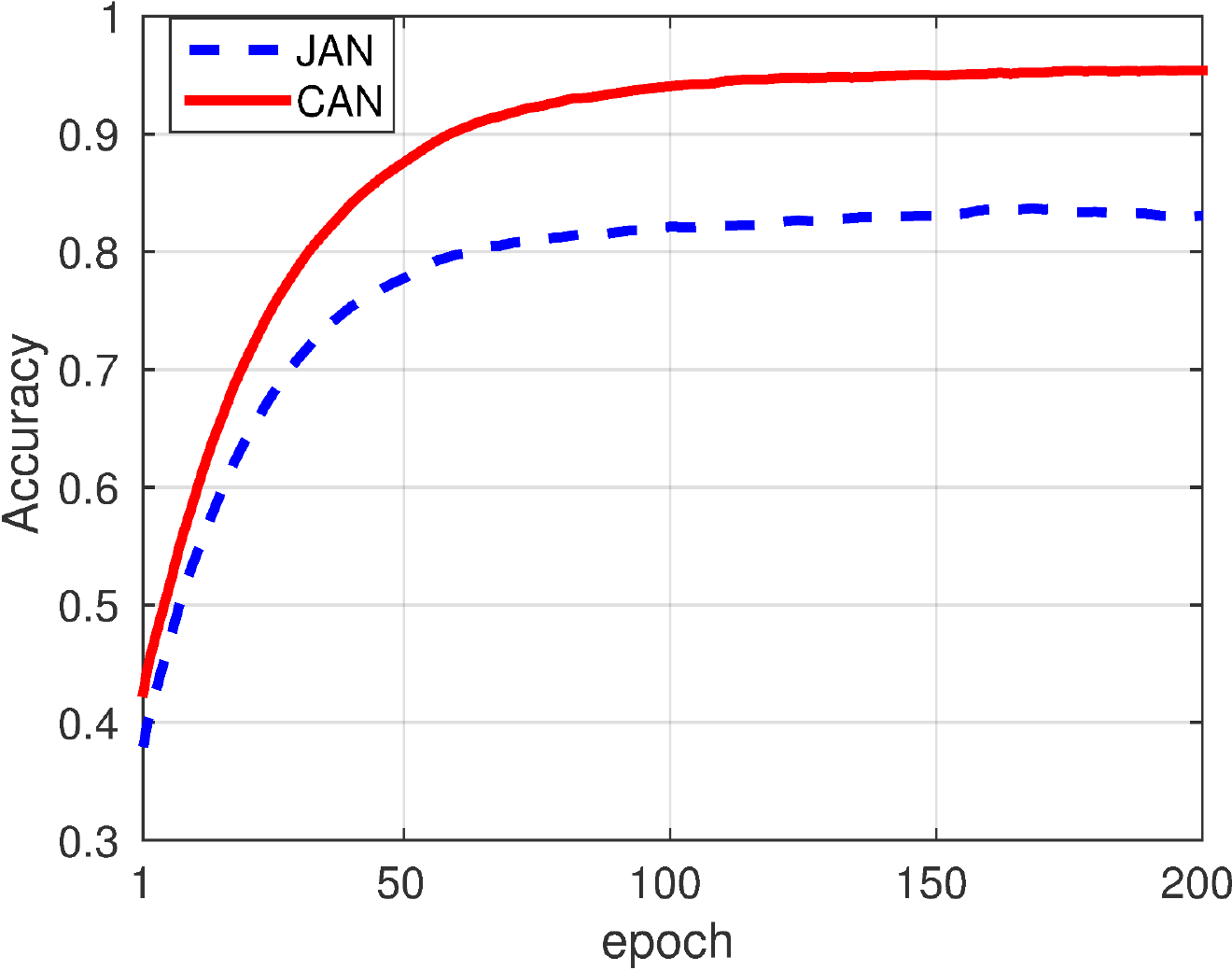}
}
\subfigure[]{
\includegraphics[viewport=100 265 500 0,scale=0.29]{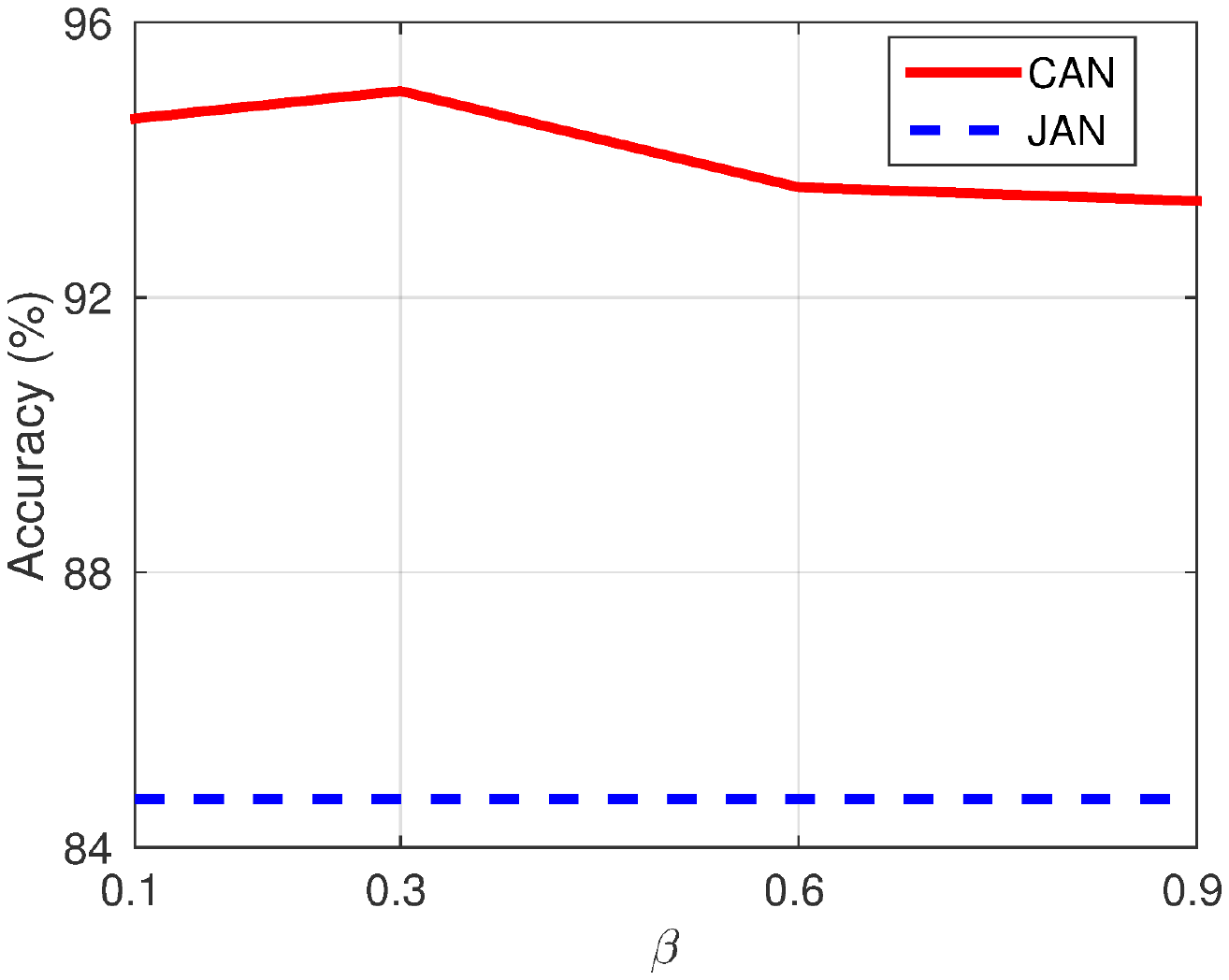}
}
\subfigure[]{
\includegraphics[viewport=110 265 500 0,scale=0.29]{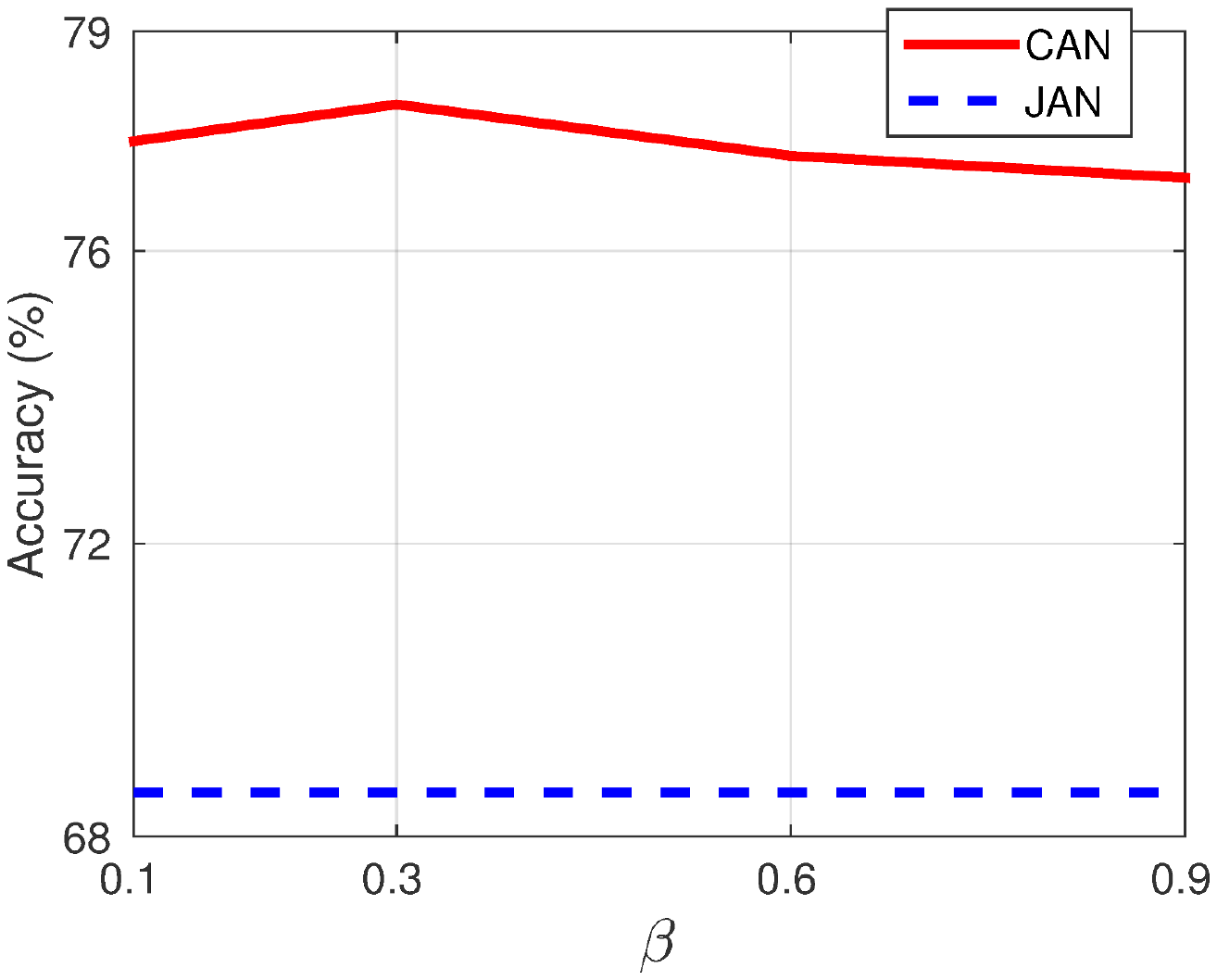}
}
\end{center}
\vspace{-3mm}
\caption{\label{fig:metric-train}
(a-b) The curve of CDD and accuracy during training on task $\textbf{A} \rightarrow \textbf{D}$ of the Office-31 dataset.
The ``CDD-G" denotes the contrastive domain discrepancy computed with ground-truth target labels.
(c-d) The sensitivity of accuracy of CAN to $\beta$.
The results for $\textbf{A} \rightarrow \textbf{D}$ (Left)
and $\textbf{D} \rightarrow \textbf{A}$ (Right)
are illustrated as examples. The trends for other tasks are similar.
}
\vspace{-3mm}
\end{figure*}

\subsection{Ablation studies} \label{sec:ablation-study}
\textbf{Effect of inter-class domain discrepancy.}
We compare our method (``CAN") with that trained using intra-class discrepancy only (``intra only"), to verify the merits of introducing inter-class domain discrepancy measure. The results are shown in the last two rows in Table \ref{table:cls-office-31} and \ref{table:cls-visda}. 
It can be seen that introducing the inter-class domain discrepancy 
improves the adaptation performance.
We believe the reason is that 
it is impossible to completely eliminate the intra-class domain discrepancy,
maximizing the inter-class domain discrepancy may alleviate
the possibility of the model overfitting to the source data and
benefits the adaptation.

\textbf{Effect of alternative optimization and class-aware sampling.}
%
Table \ref{tab:compare-optimization} examines two key components of CAN,
\ie alternative optimization (or ``AO"), and class-aware sampling (or ``CAS").
We perform ablation study by leaving-one-component-out of our framework at a time.
In Table \ref{tab:compare-optimization}, the method ``w/o. AO" 
directly employs the outputs of the network at each iteration as 
pseudo target labels to estimate CDD and back-propagates to update the network.
It can be regarded as updating the feature representations and pseudo target labels
simultaneously.
The method ``w/o. CAS" uses conventional class-agnostic sampling instead of CAS.
The comparisons to these two special cases verify the contributions of AO and CAS in our method.

Interestingly, even without 
alternative optimization, 
the method ``w/o. AO" improves over 
class-agnostic methods, 
\eg DAN, JAN, \etc.
This suggests our proposed CDD in itself is robust to the label noise to some extent,
and MMD is a suitable metric to establish CDD (see Section \ref{sec:cdd}).

\setlength{\tabcolsep}{6pt}
\begin{table}[ht]
\begin{center}
\scalebox{0.9}{
\begin{tabular}{ l  c  c c}
\toprule
Dataset     &  w/o. AO &  w/o. CAS & CAN\\
\midrule
Office-31  &  88.1 & 89.1 & 90.6 \\
VisDA-2017 &  77.5 & 81.6 & 87.2  \\
\bottomrule
\end{tabular}}
\end{center}
\caption{\label{tab:compare-optimization}
The effect of alternative optimization (AO) and CAS.
The mean accuracy over six tasks on Office-31 and 
the mean accuracy over 12 classes on VisDA-2017 validation set are reported.
}
\vspace{-2mm}
\end{table}

\setlength{\tabcolsep}{2.5pt}
\begin{table}[ht]
\small
\begin{center}
\scalebox{0.9}{
\begin{tabular}{ l  c  c   c   c   c }
\toprule
Method &  A $\rightarrow$ W & A $\rightarrow$ D & D $\rightarrow$ A & W $\rightarrow$ A & Average \\
\midrule
pseudo$_{0}$        &  85.8 & 86.3 & 74.9 & 72.3  &  79.8 \\
pseudo$_{1}$        &  90.2 $\pm$ 1.6 & 92.5 $\pm$ 0.4 & 75.7  $\pm$  0.2 & 75.3 $\pm$ 0.6  &   83.4\\
CAN        &  94.5 $\pm$ 0.3 & 95.0 $\pm$ 0.3 & 78.0  $\pm$  0.3& 77.0 $\pm$ 0.3  &  86.1 \\

\bottomrule
\end{tabular}}
\end{center}
\caption{\label{tab:compare-pseudo}
Comparison with different ways of utilizing pseudo target labels.
The ``pseudo$_0$" means training with pseudo target labels (achieved by our initial clustering) directly. The ``pseudo$_1$" is to alternatively update target labels through clustering 
and minimize the cross-entropy loss on pseudo labeled target data. In ``pseudo$_1$", 
the cross-entropy loss on source data is also minimized.
}
\vspace{-5mm}
\end{table}
\textbf{Ways of using pseudo target labels.}
The estimates for the target labels can be achieved through clustering, 
which enables various ways to train a model.
In Table \ref{tab:compare-pseudo}, we compare our method with two different ways of training
with pseudo target labels achieved by the clustering.
One way (``pseudo$_0$") is to fix these pseudo labels to train a model directly.
The other (``pseudo$_1$") is to update the pseudo target labels during training,
which is the same as CAN, but to train the model based on the cross-entropy loss
over pseudo labeled target data rather than estimating the CDD.

As shown in Table \ref{tab:compare-pseudo}, ``pseudo$_0$" leads to a model 
whose accuracy exactly matches with that of the initial clustering, 
due to the large capacity of deep neural networks.
The ``pseudo$_1$" achieves significantly better results than ``pseudo$_0$", 
but is still worse than our CAN,
which verifies that our way of explicitly modeling the class-aware domain discrepancy
makes the model better adapted and less likely to be affected by the label noise.
\textbf{CDD value during training.}
In our training, we generate target label hypothesis to estimate CDD.
We expect that the underlying metric computed with the ground-truth target labels 
would decrease steadily during training until convergence.
To do so, during training, we evaluate the \textit{ground-truth} CDD (denoted by CDD-G) for JAN and CAN with the ground-truth target labels. The trend of CDD and the test accuracy during training are plotted in Fig. \ref{fig:metric-train}. 


As we see, for JAN (the blue curve), the ground-truth CDD rapidly becomes stable at a high level
after a short decrease.
This indicates that JAN cannot minimize the contrastive domain discrepancy effectively. For CAN (the red curve), although we can only estimate the CDD using inaccurate target label hypothesis, 
its CDD value steadily decreases along training.
The result illustrates our estimation works as a good proxy of ground-truth contrastive domain discrepancy.
And from the accuracy curve illustrated in Fig. \ref{fig:metric-train}, 
we see that minimizing CDD leads to notable accuracy improvement of CAN,
compared to JAN.

\textbf{Hyper-parameter sensitivity.}
We study the sensitivity of CAN to the important balance weight $\beta$ on two example tasks $\textbf{A} \rightarrow \textbf{D}$ and $\textbf{D} \rightarrow \textbf{A}$ in Fig. \ref{fig:metric-train}. 
Generally, our model is less sensitive to the change of $\beta$. In a vast range, the performance of CAN outperforms the baseline method with a large margin (the blue dashed curve). As the $\beta$ gets larger, the accuracy steadily increases before decreasing. The bell-shaped curve illustrates the regularization effect of CDD. 

\section{Conclusion}
In this paper, we proposed Contrastive Adaptation Network to perform class-aware alignment for 
UDA.
The intra-class and inter-class domain discrepancy are explicitly modeled and optimized through end-to-end mini-batch training. 
Experiments on real-world benchmarks 
demonstrate the superiority of our model compared with the strong baselines. 

\textbf{Acknowledgement.} \small{This work was supported in part by  the  Intelligence  Advanced Research Projects Activity (IARPA) via Department of  Interior/Interior  Business  Center  (DOI/IBC)  contract number  D17PC00340.  The  U.S.  Government  is  authorized to reproduce and distribute reprints for Governmental purposes  not withstanding  any  copyright  annotation/herein. Disclaimer: The views and conclusions contained herein are those of the authors and should not be interpreted as necessarily  representing  the  official  policies  or  endorsements, either expressed or implied, of IARPA, DOI/IBC, or the U.S.Government.}

{\small
\bibliographystyle{ieee}
\bibliography{egbib}
}

\end{document}